\pdfoutput=1

\RequirePackage{snapshot}
\documentclass[11pt]{article}

\usepackage[usenames,dvipsnames,svgnames,table]{xcolor}

\usepackage{acl}

\usepackage{times}
\usepackage{latexsym}

\usepackage[T1]{fontenc}

\usepackage[utf8]{inputenc}

\usepackage{url}
\usepackage{amsmath}
\usepackage{amssymb}
\usepackage{amsthm}
\usepackage{pifont}
\usepackage{multirow}
\usepackage{hyperref}
\usepackage{graphicx}

\usepackage[capitalize]{cleveref}
\crefname{section}{\S}{\S\S}
\Crefname{section}{\S}{\S\S}
\crefformat{section}{#2\S#1#3}
\Crefformat{section}{#2\S#1#3}
\crefrangeformat{section}{\S\S#3#1#4--#5#2#6}
\Crefrangeformat{section}{\S\S#3#1#4--#5#2#6}
\crefmultiformat{section}{\S#2#1#3}{ and~\S#2#1#3}{, \S#2#1#3}{ and~\S#2#1#3}
\Crefmultiformat{section}{\S#2#1#3}{ and~\S#2#1#3}{, \S#2#1#3}{ and~\S#2#1#3}
\crefrangemultiformat{section}{\S\S#3#1#4--#5#2#6}{ and~\S\S#3#1#4--#5#2#6}{, \S\S#3#1#4--#5#2#6}{ and~\S\S#3#1#4--#5#2#6}
\Crefrangemultiformat{section}{\S\S#3#1#4--#5#2#6}{ and~\S\S#3#1#4--#5#2#6}{, \S\S#3#1#4--#5#2#6}{ and~\S\S#3#1#4--#5#2#6}

\usepackage{listings}
\usepackage{tabu}
\usepackage{booktabs}
\usepackage{inconsolata}

\usepackage{comment}

\newcommand{\ie}{\emph{i.e.}, }
\newcommand{\eg}{\emph{e.g.}, }

\newcommand{\cmark}{\ding{51}}%
\newcommand{\xmark}{\ding{55}}%

\definecolor{codegreen}{rgb}{0,0.6,0}
\definecolor{codegray}{rgb}{0.5,0.5,0.5}
\definecolor{codepurple}{rgb}{0.58,0,0.82}
\definecolor{backcolour}{rgb}{0.95,0.95,0.92}

\lstdefinestyle{mystyle}{
    backgroundcolor=\color{backcolour},   
    commentstyle=\color{codegreen},
    keywordstyle=\color{magenta},
    numberstyle=\tiny\color{codegray},
    stringstyle=\color{codepurple},
    basicstyle=\ttfamily\footnotesize,
    breakatwhitespace=false,         
    breaklines=true,                 
    captionpos=b,                    
    keepspaces=true,                 
    numbersep=5pt,                  
    showspaces=false,                
    showstringspaces=false,
    showtabs=false,                  
    tabsize=2,
}
\theoremstyle{definition}

\newcommand{\ourdata}{SMCalFlow2Text}

\title{The Whole Truth and Nothing But the Truth: \\
Faithful and Controllable Dialogue Response Generation \\
with Dataflow Transduction and Constrained Decoding}

\author{%
Hao Fang\thanks{\; Equal contribution.} \quad 
Anusha Balakrishnan$^*$ \quad 
Harsh Jhamtani$^*$\\
\quad {\bf John Bufe} 
\quad {\bf Jean Crawford} 
\quad {\bf Jayant Krishnamurthy} \\
\quad {\bf Adam Pauls} 
\quad {\bf Jason Eisner}
\quad {\bf Jacob Andreas} 
\quad {\bf Dan Klein} \\
Microsoft Semantic Machines \ \texttt{<sminfo@microsoft.com>}
}

\date{}

\begin{document}
\maketitle

\begin{abstract}
In a real-world dialogue system, generated text must
be truthful and informative while remaining fluent and adhering to a prescribed style.  
Satisfying these constraints simultaneously is difficult for the two predominant paradigms in language generation: neural language modeling and rule-based generation.  
We describe a hybrid architecture for dialogue response generation that combines the strengths of both paradigms.
The first component of this architecture is a rule-based content selection model defined using a new formal framework called \emph{dataflow transduction},
which uses declarative rules to transduce a dialogue agent's actions and their results (represented as dataflow graphs) into context-free grammars representing the space of contextually acceptable responses.
The second component is a constrained decoding procedure that uses these grammars to constrain the output of a neural language model,
which selects fluent utterances.
Our experiments show that this system outperforms both rule-based and learned approaches in human evaluations of fluency, relevance, and truthfulness.
\end{abstract}

\section{Introduction}
\label{sec:intro}

In a task-oriented dialogue system, response generation is naturally posed as a conditional language modeling problem:
dialogue agents must produce a contextually appropriate natural language string conditioned on the history of the user and agent interaction.
But unlike many language generation problems, a good dialogue response generation model is not (just) a model of typical human utterances in context. Instead, effective dialogue agents must balance fluent generation with a set of much stricter constraints.

\begin{figure}[t]
    \centering
    \includegraphics[width=\columnwidth]{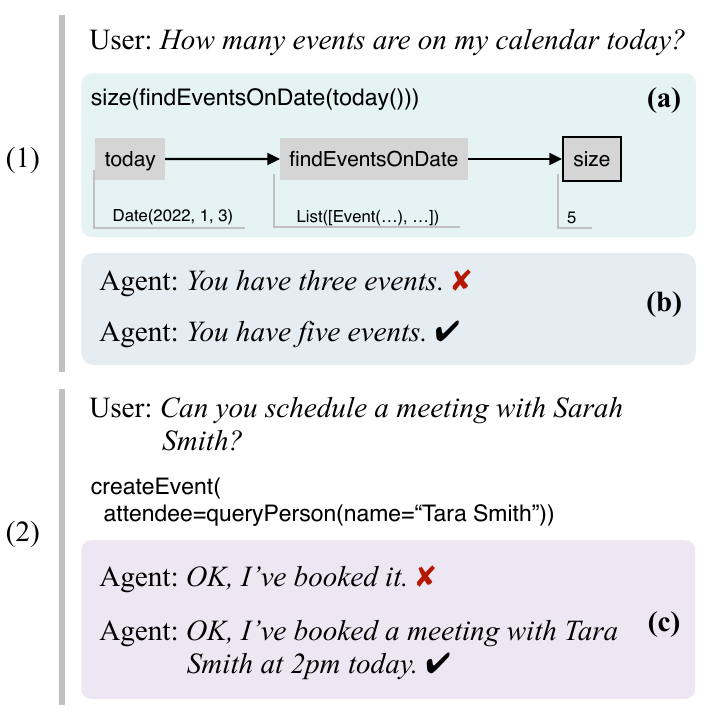}
    \vspace{-1.5em}
    \caption{
    Interaction between a user and a dialogue agent. Once the user's request is translated into an agent action---expressible as a program or dataflow graph (a)---the agent must generate a response. Agent responses might simply state the result of the agent's action, but must do so truthfully (b). Often responses should describe both the action and the result, \eg to help users identify when the agent has misunderstood their request (c). These responses should be straightforward for system designers to inspect and modify.
    \vspace{-1em}}
    \label{fig:pull}
\end{figure}

Consider the dialogue shown in \cref{fig:pull}.
In turn (1) of this dialogue, the user makes a request, which the dialogue agent correctly translates into a computation---here represented as a dataflow graph 
(\cref{fig:pull}a) in the style of \citet{SMDataflow2020}.
The agent now needs to accurately describe this computation's return value (namely, \texttt{5}).  The wrong answer in \cref{fig:pull}b shows it instead describing a different value that happens to appear elsewhere in the dataflow graph.
Turn (2) illustrates a more subtle risk: due to a speech recognition error, the agent has mistakenly created a meeting with \emph{Tara Smith} rather than \emph{Sarah Smith}.
The wrong answer in \cref{fig:pull}c shows it describing this result too briefly, which might lead the user to assume that their request was completed successfully.  To avoid confusion, a system designer might wish to ensure that the agent instead echoes back to the user the details of the agent's action.

This example highlights challenges in building real-world dialogue response generation systems.

First, response generation
is not simply a problem of describing the \emph{result} of a computation in natural language. In some cases, response generators may also usefully \textbf{describe the provenance} of that result---the computation itself and its intermediate values.
In many human-to-human conversations, a response as detailed 
as \cref{fig:pull}c would be over-informative, violating Grice's maxim of quantity~\citeyearpar{Grice1975}.  
But for a speaker that is prone to mistakes, such as an AI agent, describing its own understanding 
can increase user trust when the understanding is accurate and provides an opportunity for correction when it is not.%

Second, dialogue response generation systems must \textbf{guarantee truthfulness}: since the 
user often has no way to check the responses, even occasional errors could have disastrous consequences and would greatly undermine trust.
Yet
truthful utterances might be low-probability under a domain-general language model (LM), particularly when they reflect errors in language understanding (as in \cref{fig:pull}c).

Finally, response generation systems must \textbf{support declarative specification of agent behavior}.
When confusing or infelicitous responses are discovered, it should be possible to easily and precisely modify them without changing the dialogue agent's behavior in other contexts.

In recent years, the main focus of academic dialogue research has been on ``end-to-end'' learned models for response generation, especially neural sequence models \cite{vinyals-le-2015-neural,zhang-etal-2020-dialogpt}.
But while such models excel at producing fluent and coherent output, research continues to find that they struggle in maintaining faithfulness \cite{wiseman-etal-2017-challenges,maynez-etal-2020-faithfulness,raunak-etal-2021-curious,liu-etal-2023-evaluating,zhang-etal-2023-language}.
Perhaps more fundamentally, because the behavior of such systems is encoded implicitly in their training data, designing a dialogue system 
requires system builders to write and edit a large number of training examples
whose final effect may be difficult to predict.

As a result,
many dialogue systems in the real world remain rule-based: 
system builders hand-write rules (\eg in the form of a synchronous grammar) for transforming dialogue states into text, and these rules are applied directly during deployment.
But such rule-based systems are also notoriously difficult to build and maintain \cite{walker-etal-2002-training,reiter-2022-problems}.
They require designers to anticipate every low-level question about surface realization,
and to encode these in the same grammar that is responsible for enforcing high-level properties like truthfulness.

Given the many strengths of modern LMs, is there a way to leverage them while
satisfying the numerous other demands on dialogue response generation systems?
In this paper, we describe a hybrid approach that combines the advantages of end-to-end and rule-based approaches.
This approach has two components:

\begin{itemize}

    \item A dataflow transduction procedure (\cref{sec:dataflow_grammar})
    that maps any computation by the agent
    (represented as a dataflow graph) to a small context-free grammar (CFG) that defines the space of natural language descriptions or responses allowed for the given computation.  The mapping is defined by declarative rules.
    This formal framework makes it possible to write rules to precisely and truthfully describe both data and its provenance, while performing supplementary computation where needed to produce informative responses.\looseness=-1

    \item A constrained decoding procedure (\cref{sec:response_generation})
    that uses beam search to identify strings that are both grammatical under the CFG and probable under a given language model (LM).
    In effect, this intersects the CFG with the LM.
\end{itemize}
This makes it possible to decompose language generation into a \textbf{content selection model} (implemented by the dataflow transducer) and a separate \textbf{fluency model} (implemented by the LM).
Hybrid generation systems of this kind have a long history in NLP, dating back to 
\citet{knight-hatzivassiloglou-1995} and
\citet{langkilde-knight-1998-generation-exploits}.  They mapped an abstract meaning representation (AMR) to an acyclic finite-state automaton (FSA) and scored its paths with an $n$-gram LM.%
We replace AMR with dataflow, replace their mapping rules with dataflow transduction rules, upgrade their FSA to a CFG, and upgrade their $n$-gram LM to a neural LM.  In this way, we respectively support computation graphs, arbitrary tests and transductions, nested syntactically typed generation templates \cite[already present in][]{knight-hatzivassiloglou-1995}, and modern language models.

Together, dataflow transduction and constrained decoding allow a compact generation system to faithfully and fluently describe a complex and open-ended space of actions.
We built such a hybrid system for calendar event queries in the SMCalFlow domain \cite{SMDataflow2020}.
When evaluated on a subset of annotated dialogues,
it was consistently rated as more truthful, relevant, and fluent than either a rule-based or end-to-end neural system (\cref{ssec:main_results}).
Results were similar on MultiWOZ dialogues \cite{budzianowski-etal-2018-multiwoz,eric-etal-2020-multiwoz} (\cref{ssec:multiwoz_results}).
Code, data, and trained models used in our experiments are released
at \url{https://github.com/microsoft/dataflow2text}.

\section{Problem Formulation}

We study the problem of response generation for task-oriented dialogue.
A dialogue, like the one in \cref{fig:pull},
consists of a sequence of \textbf{turns} $k$,
each consisting of a \textbf{user utterance} $x_k$, one or more \textbf{actions} $a_k$, and
an \textbf{agent response} $y_k$.
The job of a \textbf{dialogue agent} is to predict an appropriate action and response from a dialogue history, \ie, to map from $(x_1, a_1, y_1, x_2, a_2, y_2, \ldots, x_n) \mapsto (a_n, y_n)$.\looseness=-1

How is this done?  Typically, a \textbf{language understanding module}
maps the user utterance $x_k$ (in context) 
to a \textbf{formal meaning representation}.
The agent reasons about this meaning representation to determine its own actions $a_k$.  Finally, a \textbf{response generation module} 
maps these actions or their results (in context) to the agent utterance $y_k$.\looseness=-1

The focus of this paper is the response generator.
We assume that the formal meaning representation takes the form of an executable program, as is common in the semantic parsing literature---and that the actions are produced by evaluating this program, possibly with side effects.
As described by \citet{SMDataflow2020}, the program may be viewed as a \textbf{dataflow graph} in which each node is labeled with a function, constructor, or primitive value, as well as a return value once the node is executed.
We aim to implement a response generator 
that, when applied to an evaluated dataflow graph, satisfies the three properties outlined in \cref{sec:intro}: description of data and its provenance, guaranteed truthfulness, and declarative specification.
In practice, for guidance when developing our generator, we refer to a development set of dialogues annotated with gold-standard dataflow graphs and agent responses.

\section{Dataflow Transduction}
\label{sec:dataflow_grammar}

Given a dataflow graph $G$ (\eg \cref{fig:pull}a) rooted at a node $v_\texttt{root}$
(the return value of the program represented by the dataflow graph),
our task is to generate a string that describes $v_\texttt{root}$ and its provenance.
To achieve this, we propose a new formal framework for generation based on \textbf{dataflow transduction}.
At a high level,
the formalism uses declarative rules that describe how to transform a dataflow graph into a small graph-specific grammar 
(specifically a \textbf{quasi-synchronous context-free grammar}, or QCFG)
that defines the space of allowed responses.
These rules walk along the graph, introduce new computations (dataflow subgraphs) as needed, and add rules to the grammar.

Formally, a dataflow transducer $\mathcal{S}$ is defined by a 4-tuple 
$(\mathcal{T}, \Sigma, \mathcal{R}, t_\texttt{start})$ 
where $\mathcal{T}$ is a set of nonterminal types,\footnote{%
\label{fn:nonterm}In practice, nonterminal types might correspond to dialogue acts, syntactic categories, semantic categories, etc.
This is up to the system designer.}
$\Sigma$ is the set of terminals (word types),
$\mathcal{R}$ is a set of dataflow transduction rules (see \cref{ssec:generation_rule}),
and $t_\texttt{start} \in \mathcal{T}$ is the start nonterminal.
When applied to $G$,
the dataflow transducer produces a QCFG.  As a side effect, it may extend the graph with new computations.  We use $\bar{G}$ to denote the extended graph.

A QCFG \citep{smith-eisner-2006-quasi} is a specialized CFG whose nonterminals include alignments to the nodes $V(\bar{G})$ of  $\bar{G}$. 
Where an ordinary CFG might specify ways to generate an \texttt{NP} (noun phrase) or a \texttt{DATE}, 
a QCFG would specify ways to generate an \texttt{NP} or \texttt{DATE} that describes the result and provenance of $v$, for each appropriately typed node $v \in V(\bar{G})$.
A QCFG resulting from dataflow transduction is a 4-tuple
$(\mathcal{T} \times V(\bar{G}), \Sigma, \mathcal{P}, (t_\texttt{start},v_\texttt{root}))$
where $\mathcal{T} \times V(\bar{G})$ is the QCFG's set of nonterminals 
and $\mathcal{P}$ is its set of productions.
A QCFG production has the form
    $\alpha \rightarrow \beta_1 \beta_2 \cdots \beta_N$
where the left-hand-side $\alpha = (t,v) \in \mathcal{T} \times V(\bar{G})$ is a QCFG nonterminal,
and each $\beta_i$ can be either a nonterminal $(t_i, v_i)$ or a terminal in $\Sigma$.
The $v_i$ of a right-hand-side nonterminal $\beta_i$ may have appeared in the original $G$, or may have been added to $\bar{G}$ by the dataflow transducer. These production rules then derive a set of strings just as in any CFG.

\subsection{Dataflow Transduction Rules}
\label{ssec:generation_rule}

\begin{figure}[!t]
    \centering
    \includegraphics[width=\columnwidth]{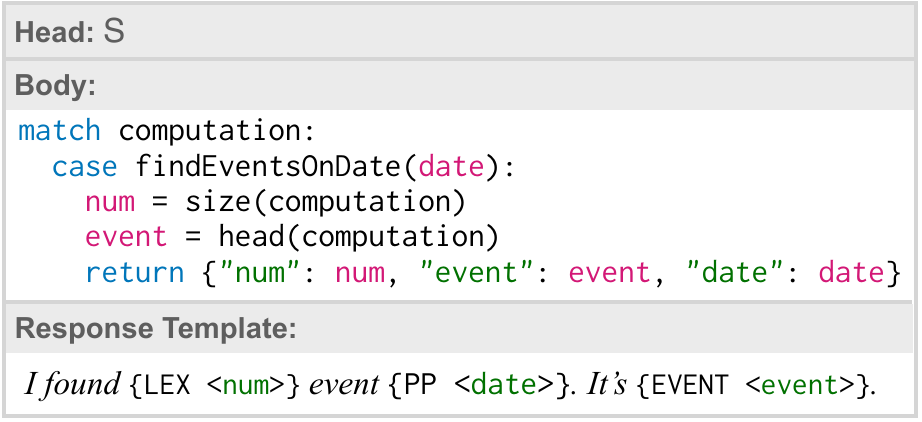}
    \caption{A dataflow transduction rule with head \texttt{S}, a body (expressed in Python), and a response template (which queries the dictionary returned by the body).}
    \label{code:genrule}
\end{figure}

A dataflow transduction rule 
is applied to a node $v \in V(\bar{G})$ (if $v$ has appropriate properties) to create a single QCFG production $(t,v) \to \cdots$ that could be used to describe $v$.
An example rule is shown in \cref{code:genrule}.
A rule has three components:
(1)~a \textbf{head}, namely the nonterminal type $t \in \mathcal{T}$;
(2)~a \textbf{body}, which is a piece of code that determines whether the rule can apply to $v$, and which may look up or create nodes that are related to $v$;
and (3)~a \textbf{response template}, which specifies the right-hand side of the QCFG production in terms of the related nodes that identified in the body.

\textbf{Rule Head.}
This nonterminal type characterizes the type of node that the transduction rule is able to describe and the type of description that it will produce.\textsuperscript{\labelcref{fn:nonterm}}
When a rule with head $t$ is successfully applied to the node $v$, the resulting QCFG production has left-hand-side $(t,v)$.

\textbf{Rule Body.}
The rule body tests whether the rule can be applied by examining the dataflow graph $\bar{G}_v$ rooted at $v$.
It also binds variables to other nodes of $\bar{G}$ that are to be described recursively.\footnote{These nodes may already exist in $\bar{G}_v$, or may represent new computations that take existing nodes of $\bar{G}_v$ as input.%
}
For example, the rule body in \cref{code:genrule} checks whether $\bar{G}_v$ has the form
\texttt{findEventsOnDate(date)}.  If so, it binds the variable \texttt{date} accordingly, and introduces new nodes into $\bar{G}$, bound to the variables \texttt{num} and \texttt{event}, which compute the number of events and the first event.
All three of these variables will be referenced in the response template.

\textbf{Response Template.}
The response template says how to create the right-hand side of the QCFG rule---a sequence $\beta_1\cdots\beta_N$ of terminals and nonterminals.  Each QCFG nonterminal $\beta_i=(t_i,v_i)$ specifies a related node $v_i \in V(\bar{G})$ to describe, along with a dataflow nonterminal $t_i$ that says \emph{how} to describe it.  The possible descriptions of $v_i$ will thus emerge from applying transducer rules with head $t_i$ to node $v_i$.
In our template syntax, the notation 
\verb+{EVENT <event>}+ would construct the QCFG nonterminal $(\texttt{EVENT}, v)$,
if the rule body has bound the variable \texttt{event} to the node $v$.
This syntax is illustrated in \cref{code:genrule}, whose response template constructs a right-hand side that intersperses terminal symbols with three QCFG nonterminals, which pair types \texttt{LEX}, \texttt{PP}, and \texttt{EVENT} with nodes that were identified by the rule body.

Our actual template format is more flexible than shown here.  It allows choices within the template in order to specify variant phrasings.\footnote{This is equivalent to specifying multiple rules with the same head and body, but more concise.}  This advanced feature is described in Appendix~\ref{appendix:response_template}.
Details and examples of dataflow transduction rules used in our experiments are provided in Appendix~\ref{appendix:dataflow_transduction_rules}.

\begin{figure*}[!t]
    \centering
    \includegraphics[width=.98\textwidth,clip,trim=2.2in 1.6in 1.4in 0.6in]{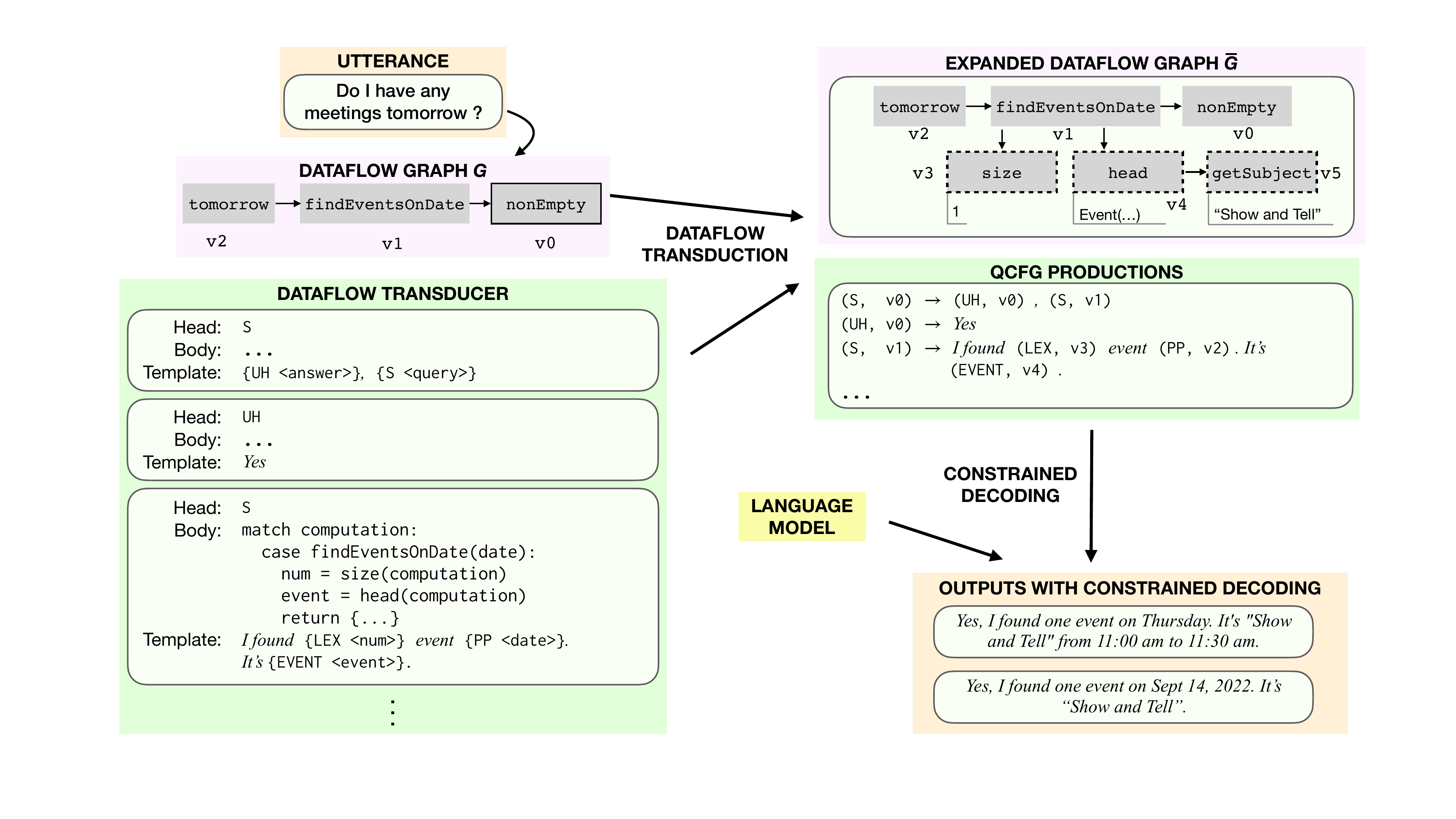}
    \caption{%
        The hybrid response generation approach using dataflow transduction and constrained decoding.
        Given a computation \texttt{nonEmpty(findEventsOnDate(tomorrow()))} for the user utterance 
        \textit{``Do I have any meetings tomorrow''},
        we first derive QCFG productions by applying the dataflow transducer to the dataflow graph $G$ using 
        the procedure described in \cref{ssec:qcfg_productions}.  This procedure also expands the dataflow graph into $\bar{G}$: for example, the nodes \texttt{v3} and \texttt{v4} were added by the third transducer rule.
        Then we extract candidate responses from a LM, constrained by the QCFG.
        The varying descriptions of the date \texttt{v2} and the event \texttt{v4} 
        are permitted because the QCFG offers a choice of productions that can be used to expand the $(\texttt{PP},\texttt{v2})$ and $(\texttt{EVENT},\texttt{v4})$ nonterminals.
        (Those productions and the transducer rules that created them are not shown in the figure. The nodes added by those transducer rules and used by those productions are also not shown, except for \texttt{v5}.)}
    \label{fig:system_pipeline}
\end{figure*}

\subsection{Dataflow Transduction Procedure}
\label{ssec:qcfg_productions}

Given a dataflow transducer $\mathcal{S}$ and a dataflow graph $G$ rooted at node $v_\texttt{root}$,
we can transduce the graph into a QCFG as follows.
The system starts out by creating QCFG productions that can expand the start nonterminal $(t_\texttt{start}, v_\texttt{root})$.
For each transduction rule in $\mathcal{R}$ whose head is $t_\texttt{start}$, 
it executes the body, which checks any additional conditions for whether the rule can be applied to $v_\texttt{root}$,
binds variables, and uses the response template to create a QCFG production.
If these productions mention new nonterminals,
the system recursively creates further QCFG productions, 
in the same way, that can expand those nonterminals.
As a special case, to expand a nonterminal of the form
$(\texttt{LEX},v)$, the system creates a QCFG production whose
right-hand side gives the value of $v$, as rendered
into natural language using a lexicalization function rather than a template; 
\eg a value \texttt{Integer(42)} would be rendered as \textit{``42''}. 

The recursive process continues until productions have been created for every nonterminal that appears in the QCFG.
The resulting QCFG compactly represents a combinatorial space of
possible responses.  It will generally include multiple productions aligned to the
same node $v$, created by different dataflow transduction rules.

This mechanism can be used to copy simple values like strings and numbers from the dataflow graph,
as well as to create more complex recursive descriptions.
Note that (1) transduction rules are selected via their head but also condition on the dataflow graph through their body,
and (2) all QCFG nonterminals are grounded in the dataflow graph. 
Together, this provides a means to ensure truthfulness when generating responses.

Note there may be multiple transduction rules for each QCFG nonterminal $\beta_i$
and the QCFG may admit combinatorially many derivation trees.
Each of these derivation trees derives a truthful response. 
However, since different trees use different rules, the responses may vary in their information content, 
presentation order, linguistic style, and choice of terminals.
The amount of variation can be controlled by the author of the dataflow transducer.
In this paper, we use a neural LM with constrained decoding to select a fluent and appropriate response from all these truthful responses, as
described in the next section (\cref{sec:response_generation}).
\section{Constrained Decoding}
\label{sec:response_generation}\label{sec:constrained_decoding}

In this section, we describe how to integrate the formal framework above with a general LM to perform response generation, as illustrated in \cref{fig:system_pipeline}.
Given a derived QCFG of the kind
described in \S\ref{ssec:qcfg_productions},
we will perform constrained decoding as in \cite{shin-etal-2021-constrained,subhro-etal-2022-benchclamp}, 
generating response candidates from a pretrained LM.

The QCFG resulting from dataflow transduction implicitly represents a set of possible derivation trees and the agent responses they yield. As long as
transduction rules faithfully describe the nodes they apply to, every derivation in this set will correspond to a truthful agent utterance.
But these utterances may not always be grammatical or natural.
For example, the response template in \cref{code:genrule} may be realized as 
\textit{``I found 2 event on Monday''} since the rule body does not check whether the value of \texttt{num} is 1. 
Similarly, the response template
\begin{center}
    $\bigl\{$\texttt{EVENT} $\langle$\textit{event}$\rangle$$\bigr\}$ starts on $\bigl\{$\texttt{DATE} $\langle$\textit{date}$\rangle$$\bigr\}$.
\end{center}
may be realized as
\textit{The product meeting on Monday starts on Monday}, if the grammar permits identifying events by their dates.
With carefully engineered and highly specialized rules (\eg using extremely fine-grained nonterminal types),
it would be possible to ensure that the responses are always fluent and even that there is always a single possible outcome from the top-down search procedure.
However, this would usually require much a more complicated set of rules, which creates a burden for system development and maintenance.

Our proposed approach instead uses a large-scale pretrained LM (preferably fine-tuned) 
to select among truthful utterances produced by the QCFG.\footnote{Of course, decisions deferred to the LM could be encoded in the grammar  instead. While this is rarely necessary to ensure grammaticality or fluency, system designers might choose to encode some \emph{pragmatic} decisions, like how much detail to provide, in the grammar rather than in the LM.}
One option is to use the LM to re-rank all strings that can be produced by the QCFG, but that would be very computationally expensive even when that set is finite.
Instead, we follow \citet{shin-etal-2021-constrained} and \citet{subhro-etal-2022-benchclamp},
who decode sentences from a given LM under the constraint that they must be valid under a given CFG.
In contrast to these prior papers, which used a static CFG, we derive a new CFG each time the dialogue agent needs to generate a response, by applying the dataflow transducer to the current dataflow graph.

The constrained decoding process is a special case of beam search.  For each $\ell = 0, 1, \ldots$, it maintains up to $K$ prefixes of the same length $\ell$ and tries to extend each in all legal ways to length $\ell+1$, pruning back to the $K$ most probable extensions.  For each prefix $y_1 y_2 \ldots y_\ell$ and each terminal symbol $y_{\ell+1} \in \mathcal{T}$, the extension $y_1 y_2 \ldots y_{\ell+1}$ is only legal if it is a prefix of some legal complete response---i.e., some string that is grammatical under the QCFG.  This check can be efficiently performed via an incremental context-free parsing algorithm \cite{earley70}
using the parsing state of the prefix $y_1 y_2 \ldots y_\ell$.
In other words, constrained decoding only considers a prefix if it could be extended into at least one legal complete response.
Note that the combinatorially many legal responses are never enumerated individually.
Rather, the set is compactly represented by the set of QCFG productions.

\section{Experiments}

To evaluate the proposed approach, we conducted a set of detailed experiments (\cref{ssec:data}--\cref{ssec:ablation}) on a subset of the
SMCalFlow dataset \cite{SMDataflow2020},
and a brief study (\cref{ssec:multiwoz_results}) applying our approach to the MultiWOZ dataset \cite{budzianowski-etal-2018-multiwoz}.

\begin{table*}[t]
    \small\hspace{-1ex}%
    \begin{tabular}{lcccccccc}
        \toprule

        \multirow{2}{*}{\textbf{System}} 
        & \multicolumn{5}{c}{\textbf{Automatic Metrics}} 
        & \multicolumn{3}{c}{\textbf{Human Evaluation (\%)}} \\
        \cmidrule{2-9}

        & \textbf{BLEU} 
        & \textbf{ROUGE} 
        & \textbf{BERTSc.} 
        & \textbf{R@1} 
        & \textbf{R@5}
        & \textbf{Grammatical} & \textbf{Relevant} & \textbf{Truthful} \\
        \midrule

        QCFG Random Sampling
        & $.35$  & $.58$ & $.50$ & $.02$ & $.06$ 
        & $62.3$ & $90.9$ & ${\bf 92.3}$ \\

        Unconstrained Decoding
        & $.77$ & $.87$ & $.87$ & $.47$ & $.66$
        & ${\bf 98.7}$ & ${\bf 93.3}$ & $82.2$ \\

        QCFG-Constrained Decoding
        & $.80$ & $.86$	& $.85$ & $.56$	& $.78$
        & ${\bf 99.0}$ & ${\bf 96.6}$ & ${\bf 91.6}$ \\ 

        \midrule

        Gold 
        & $1.0$ & $1.0$ & $1.0$ & $1.0$ & $1.0$
        & ${\bf 99.0}$ & ${\bf 98.0}$ & ${\bf 92.3}$ \\

        \bottomrule
    \end{tabular}
    \caption{Evaluation results on \ourdata{}.
        Automatic metrics are calculated against the gold responses on the full validation set.
        Human evaluation is conducted on 297 randomly sampled validation examples.
        We boldface the best result in each Human Evaluation column, along with results that are not significantly worse ($p < 10^{-4}$, McNemar's test).
    }
    \label{tab:calflow_main_results}
\end{table*}

\subsection{Data and Evaluation Metrics}
\label{ssec:data}

SMCalFlow is a large-scale task-oriented dialogue dataset, in which each user utterance is annotated with a correct dataflow program 
(i.e., computation) 
and a ``gold'' response that would be desirable for the agent to produce.\footnote{%
\label{fn:gold}The ``gold'' responses were generated from an earlier system, but were manually validated by human experts.  Like ours, the earlier system also contained rules and constraints.}
We use the v2.0 release processed by \citet{platanios-etal-2021-value}.
We focus on a subset of SMCalFlow involving calendar event queries.
This subset contains 8938 training examples and 1041 validation examples.
We found that 187 transduction rules, written by some of us in a matter of hours,
were sufficient to cover all gold system responses in these examples.\footnote{Some of our rule bodies chose to expand the dataflow graph by calling functions, so we also had to implement those functions.  In an end-to-end dialogue system, most of those functions would already have been implemented to support agent actions, not just natural language responses.}
We package the annotated examples, transduction rules, and necessary meta information for executing the dataflow programs
as a new dataset, \ourdata{}.

\textbf{Automatic Metrics.}
For automatic evaluation, we use several reference-based metrics:
BLEU-4 \cite{papineni-etal-2002-bleu} and ROUGE-L \cite{lin-2004-rouge}
are computed using GEM-metrics,\footnote{\url{https://github.com/GEM-benchmark/GEM-metrics}}
and BERTScore-F1 is computed using HuggingFace Evaluate.\footnote{\url{https://github.com/huggingface/evaluate}}
Following the recommendation of \citet{DBLP:conf/iclr/ZhangKWWA20}, we use the re-scaled version of BERTScore, which is easier to interpret.
We additionally consider exact match scores, 
i.e., \textbf{R@K}, which measure whether one of the top $K$ response candidates 
exactly matches the reference.
Both \textbf{R@1} and \textbf{R@5} scores are reported.
We lowercase all the strings and remove any extra spaces in the predictions and references 
before computing the evaluation metrics.

\textbf{Human Evaluation.}
It is well-known that popular automatic evaluation metrics may not always reflect the true quality of the generated responses \cite{celikyilmaz-etal-2021-evaluation}.
Thus, we further carry out human evaluation on 297 examples randomly sampled from the validation data.
Specifically, for each generated response, we collect human judgments on three questions:
\textbf{grammaticality} (\textit{``has the virtual assistant made any grammar errors?''}),
\textbf{relevance} (\textit{``has the virtual assistant misunderstood the user’s request?''}),
and \textbf{truthfulness} (\textit{``has the virtual assistant provided any incorrect information as judged using the database and timestamp?''}).
Three judgments are collected for each question,
and we report the percentage of examples where \textit{``no''} is the majority-voted answer.  Higher percentages are better.
Crowdworkers are recruited from Amazon Mechanical Turk with qualification
requirements such as having a work approval rate higher than $80\%$ and having performed a minimum of $100$ annotations. 
They are paid at the rate of \$$0.15$ per judgment.
For responses generated by the constrained decoding approach, 
annotators generally agree with each other on the three questions, \ie the percentage of examples where all three workers choose the same answer
are around $90\%$, $78\%$ and $76\%$, respectively.
More details are provided in \cref{appendix:human_eval}.

\subsection{Main Results}
\label{ssec:main_results}

\autoref{tab:calflow_main_results} shows our main evaluation results on \ourdata{}.
The first baseline we considered is to randomly sample responses from the generated QCFG.
The other baseline is unconstrained LM decoding without using dataflow transduction.
Model outputs are compared to ``gold'' agent utterances.
For both unconstrained and constrained decoding, 
the text used to prompt the LM is 
a string representation of the computation graph
(in the format released in SMCalFlow v2.0),
followed by its execution result rendered as a JSON string.
In both cases,
we decode using beam search with beam size $K=5$.
The LM is initialized from CodeT5-base \cite{wang-etal-2021-codet5} and
fine-tuned on all training examples.
See \cref{appendix:hyperparams} for more details.

As expected, the QCFG random sampling baseline struggles on all the automatic metrics, 
since dataflow transduction rules are written with an emphasis on truthfulness rather than fluency. 
This is reflected in the grammaticality score from the human evaluation as well.
However, the truthfulness score matches that of the gold responses ($92.3\%$): this baseline rarely generates incorrect responses.  Its responses are sometimes generic and omit information that would be relevant to the user---its relevance score is the lowest among all compared approaches---although this behavior contributes to the high truthfulness score.

In contrast, unconstrained decoding LM without dataflow transduction achieves impressive scores on automatic evaluation.
Human evaluation also suggests that the generated responses are grammatically correct and relevant to the user's request in most cases.
However, unconstrained decoding scores low on truthfulness, making false statements in about one-fifth of the generated responses.
This high rate of factual errors from neural LMs is consistent with findings in prior work \cite{wiseman-etal-2017-challenges,maynez-etal-2020-faithfulness,raunak-etal-2021-curious,liu-etal-2023-evaluating,zhang-etal-2023-language}.
It is usually unacceptable in real-world applications.

\begin{table*}[t]
    \centering
    \footnotesize
    \begin{tabular}{c|p{1.3in}|p{1.2in}|p{1.2in}|p{1.4in}}
    \toprule
    & {\bf User Utterance} 
    & {\bf Gold} 
    & {\bf QCFG-Constrained Decoding} 
    & {\bf Unconstrained Decoding} \\

    \midrule
    A. 
    & {\it Do I have any events scheduled on Sunday the 3rd?}
    & {\it No, I didn't find any event on March 3rd.} 
    & {\it No, I didn't find any event on March 3rd.} 
    & {\it No, I didn't find any event on {\color{red}May} 3rd.} 
    \\

    \midrule
    B. 
    & {\it Does Pikachu fly in tomorrow?}
    & {\it Yes, I found one event matching ``flying in'' with Pikachu tomorrow. It's ``fly in'' from 4:00 to 4:30 PM.} 
    & {\it Yes, I found one event matching ``flying in'' with Pikachu tomorrow. It's from 4:00 to 4:30 PM.} 
    & {\it Yes, I found one event matching ``flying in'' with Pikachu tomorrow. It's ``Fly in'' from {\color{red}5:00 to 5:30} PM.} 
    \\

    \midrule
    C. 
    & {\it Can you find me any events during last March?}
    & {\it Yes, I found one event last March. It was ``Dancing in Athenny'' on March 17, 2019 from 4:00 to 4:30 PM.} 
    & {\it Looks like it.} 
    & {\it Yes, I found one event last March. It was ``{\color{red}Erin Go Bragh}'' on March 17, 2019 from {\color{red}5:00 to 5:30} PM.} 
    \\

    \bottomrule
    \end{tabular}
    \caption{
    Example predictions from QCFG-constrained decoding and with unconstrained decoding. 
    In all the examples shown, outputs from unconstrained decoding are untruthful to the database due to content hallucination even though the model has access to the correct execution results as part of the prompt.
    We observe that in a few cases, the constrained model prefers truthful but pragmatically unhelpful omissions like such as \textit{``Looks like it''}
    (in Example C) compared to a more specific response.
    }
    \label{tab:examples}
\end{table*}

Compared with unconstrained decoding, our proposed QCFG-constrained decoding 
achieves significantly better scores on exact match, truthfulness, and
even relevance, while maintaining similar scores on BLEU, ROUGE, BERTScore and grammaticality.
In particular, human evaluation results indicate that the quality of generated responses 
is very close to that of the gold responses.
\cref{tab:examples} shows some example predictions.
We share some qualitative analysis in \cref{appendix:qualitative_analysis}.

Since even the gold responses did not achieve 100\% on human evaluation scores, we manually inspected
those problematic examples.
There are 4 examples for which the majority-voted answer to the ungrammaticality question 
is \textit{``yes but understandable,''}
and others are all rated as not containing any grammar errors.
For the relevance question, 4 examples are due to arguably bad data and 2 examples 
receive tied votes.
For the truthfulness question, 9 examples are due to arguably bad data, 
8 examples are due to to crowd worker mistakes, 
and 6 examples receive tied votes.

\subsection{Ablation Study}
\label{ssec:ablation}

\begin{table}[t]
    \centering
    \small
    \begin{tabular}{cccccc}
        \toprule

        & \textbf{BLEU} 
        & \textbf{ROUGE} 
        & \textbf{BERTSc.} 
        & \textbf{R@1} 
        & \textbf{R@5} \\

        \midrule
        \multicolumn{6}{c}{\textbf{1. LM without fine-tuning}} \\
        \midrule

        \xmark
        & $.00$ & $.03$ & $-.47$ & $.00$ & $.00$ \\
        \cmark 
        & $.04$ & $.28$ & $.05$ & $.02$ & $.02$  \\

        \midrule
        \multicolumn{6}{c}{\textbf{2. LM fine-tuned on 3\% training data}} \\
        \midrule

        \xmark
        & $.68$ & $.81$	& $.80$	& $.26$	& $.40$ \\
        \cmark
        & $.73$	& $.83$	& $.80$	& $.39$ & $.62$ \\

        \midrule
        \multicolumn{6}{c}{\textbf{3. LM fine-tuned on full training data}} \\
        \midrule

        \xmark
        & $.77$	& $.87$	& $.87$	& $.47$	& $.66$ \\
        \cmark
        & $.80$ & $.86$	& $.85$ & $.56$	& $.78$ \\

        \midrule
        \multicolumn{6}{c}{\textbf{4. LM prompted without execution results}} \\
        \midrule

        \xmark
        & $.58$	& $.70$	& $.72$ & $.27$ & $.42$ \\
        \cmark 
        & $.78$	& $.85$ & $.84$ & $.54$ & $.77$ \\

        \midrule
        \multicolumn{6}{c}{\textbf{5. LM prompted with user utterance}} \\
        \midrule

        \xmark
        & $.77$ & $.87$	& $.87$	& $.48$	& $.65$ \\
        \cmark
        & $.79$ & $.86$	& $.84$	& $.57$	& $.78$ \\

        \bottomrule
    \end{tabular}
    \caption{SMCalFlow ablation results, varying the amount of fine-tuning data (groups 1--3) and the context used in the prompt (groups 4--5).  \xmark\ and \cmark\ on the first column use
    unconstrained and QCFG-constrained decoding, respectively.}
    \label{tab:calflow_ablation}
\end{table}

We next analyze how
the amount of fine-tuning data and the context used in the prompt
impact the quality of generated responses.
Results are summarized in \autoref{tab:calflow_ablation}.

\noindent
\textbf{Impact of fine-tuning}:
Without fine-tuning the LM, neither unconstrained nor constrained decoding works well.
This is likely due to the mismatch between the pre-training tasks and the response generation task.
However, after fine-tuning on only a random 3\% of the training data, both approaches achieve significantly 
better scores, with larger gains on QCFG-constrained decoding.
This suggests that QCFG-constrained decoding is much more data-efficient in the low-data regime (268 examples).
Indeed, QCFG-constrained decoding using 3\% of the training data is on par with unconstrained decoding using 100\% of the training data, indicating 
that several expert hours spent on creating dataflow transduction rules
hold almost equivalent value to a large volume of training data.
While gaps between unconstrained and QCFG-constrained decoding
on automated metrics are small in the full-data setting  (\autoref{tab:calflow_main_results}),
unconstrained decoding still performs poorly on the truthfulness evaluation.
Thus, truthfulness failures from unconstrained decoding are not straightforwardly solved by scaling up training data;
QCFG-constrained decoding offers an easier path to
faithful response generation.

\noindent
\textbf{Impact of context}:
Results in groups 3--5 in \autoref{tab:calflow_ablation} all use 100\% of the training examples to fine-tune the LM.
The difference is in the context used in the LM prompt (during both training and inference).
For group 3, the text used to prompt the LM is the computation concatenated with the execution result, which is the same setup used 
in \cref{ssec:main_results}.
For group 4, we omit the execution results from the LM prompt (but not from the decoder constraints),
whereas for group 5, we add the user utterance (prefixed to the computation).
Comparing group 3 and group 4, omitting execution results significantly 
harms the performance of unconstrained decoding. 
In contrast, dataflow transduction rules can execute the computation internally, and do not require the LM to condition on it.
Comparing group 3 and group 5, adding the user utterance to the LM prompt does not benefit both approaches much.

\subsection{Experiments with MultiWOZ Dataset}
\label{ssec:multiwoz_results}

To demonstrate the general applicability of our approach for response generation,
we carry out a brief study on
the widely used MultiWOZ 2.1 dataset \cite{budzianowski-etal-2018-multiwoz,eric-etal-2020-multiwoz}.
We automatically convert the agent action annotations to dataflow computations
and write 14 transduction rules.
For generating responses, we use the predicted agent actions from the MTTOD system \cite{lee-2021-improving-end}.
Similar to our experiments on SMCalFlow, we fine-tune CodeT5-base on all training examples, using 
the ground-truth belief state and predicted action as the text used to prompt the LM.
For evaluation, we randomly sample 100 examples from the test split,
and two authors manually rate the generated responses from our QCFG-constrained decoding system and the MTTOD system.
The inter-annotator agreement is 100\%.
Almost all generated responses are grammatically correct and relevant to the user utterance.
To rate truthfulness, we use the predicted actions as the references.
Our QCFG-constrained decoding approach produces truthful responses for all 100 examples, 
whereas only 89 responses from the MTTOD system are truthful with respect to the predicted actions.
Among the 11 remaining examples, 7 of them are due to imperfect delexicalization and 4 are due to hallucination.

\section{Related Work}

One line of response generation research focuses on generating fluent and coherent responses 
directly from user utterances without any intermediate structured representation.
This paradigm is mostly used for chatbots,
as in early rule-based systems \cite{weizenbaum-1966-eliza,wallace-etal-2009-alice},
neural conversation models \cite{vinyals-le-2015-neural,shang-etal-2015-neural,sordoni-etal-2015-neural,li-etal-2016-deep,Serban2016AAAI},
and recent large-scale pretrained LMs like DialoGPT \cite{zhang-etal-2020-dialogpt} and GPT-3 \cite{brown-etal-2020-gpt3}.

Another line focuses on generating text from structured data, with applications 
beyond dialogue response generation.
For example, the WebNLG challenge \cite{gardent-etal-2017-webnlg} generates natural language descriptions 
from relation tuples,
and \citet{lebret-etal-2016-neural} generate a biography from a structured ``infobox'' record.
Many recent dialogue response generation tasks adopt dialogue-act-based meaning representations, 
including the MultiWOZ dataset \cite{budzianowski-etal-2018-multiwoz},
the Schema-Guided dialogue dataset \cite{rastogi2020towards},
and the E2E NLG challenge \cite{DBLP:journals/csl/DusekNR20}.
In contrast, our response generation task uses computations as the input,
which do not directly encode the dialogue acts of the responses.
This is a more challenging task, as the system needs to perform extra reasoning to obtain the derived information.
In this sense, our task is similar to the one in CoSQL \cite{yu-etal-2019-cosql}
and Logic2Text \cite{chen-etal-2020-logic2text}.

Constrained decoding techniques for neural LMs have been developed for text generation 
with different types of constraints \cite{balakrishnan-etal-2019-constrained,DBLP:conf/iclr/DathathriMLHFMY20,lu-etal-2021-neurologic,lu-etal-2022-neurologic}.
As \cref{sec:constrained_decoding} noted, we follow \citet{shin-etal-2021-constrained} but choose our grammar constraints dynamically for each response.

\section{Conclusion}

We have described a hybrid approach for building dialogue response generation systems. 
Our approach introduces a new formalism for transducing a dataflow graph into a QCFG,
which is then used in a constrained decoder that intersects the QCFG with a neural LM.
This formal framework makes it possible to write rules to precisely and 
truthfully describe data and its provenance while deferring surface realization decisions to a flexible language model.

This new approach outperforms 
unconstrained conditional language modeling in both automatic and human evaluations,
especially on truthfulness.
Moreover, using 3\% of the training data, the constrained decoding approach is on par with the unconstrained 
decoding approach when it uses 100\% of the training data, 
indicating that several expert hours spent on authoring rules hold almost equivalent value to a large volume of training data.

\section{Limitations and Future Directions}
Authoring transduction rules is relatively easy but may still be labor-intensive for complex domains.
Future work might explore (semi-)automatically deriving transduction rules from data, 
learning to synthesize them from domain specifications, 
or curating a collection of domain-general transduction rules that can be imported into new domains.

Our experiments in this paper generated text only in English.
It would be interesting to apply the framework to datasets in other languages, \eg GlobalWoZ \cite{ding-etal-2022-globalwoz}.
While our framework is intended to be agnostic to the output language, our notation for response templates might need to be slightly extended (along the lines of \cref{appendix:response_template})
to be more convenient to use with morphologically complex languages or free-word-order languages.  In these settings, presumably, the QCFG should systematically  generate many inflections or orderings for the LM to choose among.

To support multilingual dialogue systems, future work could consider automatically translating the response templates into additional languages---perhaps by working backwards from automatic translations of natural language responses that use those templates.  

Relatedly, we have only tested the proposed approach on dataflow graphs.
Future work could apply the method to generate textual descriptions of other graph-structured inputs, such as graph databases or abstract meaning representation (AMR) graphs.

While QCFG productions were unweighted in this paper, giving them weights would allow the QCFG to express its own preferences about which productions to use for a given input.  For example, in a product-of-experts architecture, the probability of a given response $y$, 
would be proportional to the LM probability of $y$ times the weights of all productions used in the QCFG derivation of $y$ (summed over all such derivations).  Beam search (\cref{sec:constrained_decoding}) could then be carried out using prefix weights \cite{opedal-et-al-2023}.  The weights could be trained using gold responses.

Weighting the QCFG raises the possibility that the dataflow transduction rules could encode pragmatic context-dependent policies.  For example, a dataflow transduction rule could call a neural network to assess the suitability of applying the rule to a given node in the dataflow graph, and then weight the resulting QCFG production accordingly.  

\section*{Ethics Statement}

Our proposed approach
strongly outperforms a purely neural model at truthfully describing the result of a computation and its provenance. However, our approach can still make pragmatically unhelpful omissions, making it potentially risky to deploy in some scenarios. 
Additionally, we leverage pre-trained neural language models such as CodeT5, and as such, we acknowledge that our approach might inherit some biases present in these pre-trained models. 

\section*{Acknowledgements}
We would like to thank Ben Van Durme, Baolin Peng, Subhro Roy, 
Richard Shin, and Patrick Xia for valuable discussions and feedback.
We also thank the anonymous reviewers for their insightful comments and suggestions.

\bibliography{main}

\clearpage
\appendix
\setcounter{table}{0}
\renewcommand{\thetable}{A\arabic{table}}

\setcounter{figure}{0}
\renewcommand{\thefigure}{A\arabic{figure}}

\section{Alternatives in Response Templates}
\label{appendix:response_template}

A dataflow transduction rule can be equipped with multiple templates, and our template format also allows choices within a single template.
Specifically, our implementation allows the use of vertical bar to encode alternatives within a template, 
\eg ``\textit{I \{\{ didn't find any | found no \}\} \{\{ matching events | events matching \{LEX [subject]\} \}\} on your calendar.}''
During dataflow transduction, this template is automatically converted into a small system of QCFG productions, 
\ie introducing new nonterminals for the alternations.

\section{Dataflow Transduction Rule Details}
\label{appendix:dataflow_transduction_rules}

In our experiments, 
there are 9 head types (including the START symbol) for the 187 transduction rules for \ourdata{},
and 3 head types for the 14 transduction rules for MultiWOZ.
Our framework is agnostic to the nonterminal types (see footnote~\labelcref{fn:nonterm}).
We mainly used syntactic categories like \texttt{NP}, \texttt{PP}, \texttt{DT}, \texttt{VB}, \texttt{UH} (interjection),
\texttt{Copula}, etc.
One potential challenge is that the domain developers may need to have some linguistic knowledge about the syntactic categories.
Alternatively, they could use semantic categories.

The complete set of rules for \ourdata{} is available in our released Python code.
The 187 transduction rules cover the 8938 and 1041 examples from the training and validation set in the original SMCalFlow data,
\ie the gold agent responses can all be produced from the transduction rules.
The authors
who wrote the rules were able to look at both the training and validation examples.
The remaining training and validation examples in the original SMCalFlow dataset are not covered by these rules.

Below we explain some examples of dataflow transduction rules.

\begin{lstlisting}[language=Python,morekeywords={match,case},frame=single]
# Head: PP
# Body:
match computation:
  case FullMonthofPreviousMonth(month):
    return {"month": month}
# Response Template:
"last {NP [month]}"
\end{lstlisting}
The rule head \texttt{PP} suggests that the computation is described as a preposition phrase.
The body simply checks whether the computation being described is a call to the function \texttt{FullMonthofPreviousMonth}
and extracts the argument \texttt{month}.
The response template lexicalizes the function call as ``last'' and defers describing the month to appropriate \texttt{NP} rules such as the one below.

\begin{lstlisting}[language=Python,frame=single]
# Head: NP
# Body:
if computation.__value__ == Month.March:
    return {}
# Response Template:
"March"
\end{lstlisting}
For this rule, its body checks the value of the computation rather than its structure.
Since the response template has no nonterminal, the body does not return any variable binding.
Note returning an empty dictionary is different from returning \texttt{None} (which is the default return value in Python),
as the latter indicates that the rule cannot be applied.

\begin{lstlisting}[language=Python,morekeywords={match,case,as},frame=single]
# Head: S
# Body:
match computation:
  case GetAttr(
    StructAttribute("organizer", _),
    event,
  ) as organizer:
    return {"event": event, "organizer": organizer}
# Response Template:
"{NP [organizer]} {{ is | are }} the {{ organizer | organizers }} of {NP [event]}".
\end{lstlisting}
The head of this rule is \texttt{S}, which is our start nonterminal.
The function \texttt{GetAttr} is similar to Python's builtin \texttt{getattr} method,
\ie it is used to access the values of an object's attributes,
and the special constructor
\texttt{StructAttribute} specifies the name of the attribute and optionally its type.
Here, the body checks whether the computation is describing the organizer of an event,
as reflected in the response template as well.
Note the response template uses the vertical bar for alternatives, as described in \cref{appendix:response_template}.
A more precise rule could choose between \texttt{are} and \texttt{is} based on whether there are multiple organizers or not.
We usually recommend leaving such decisions to the neural LM instead of hard-coding them in transduction rules,
but the latter approach is still possible if the system designer prefers. 
\section{Human Evaluation Details}
\label{appendix:human_eval}

\begin{figure*}[!t]
    \centering
    \includegraphics[width=\textwidth]{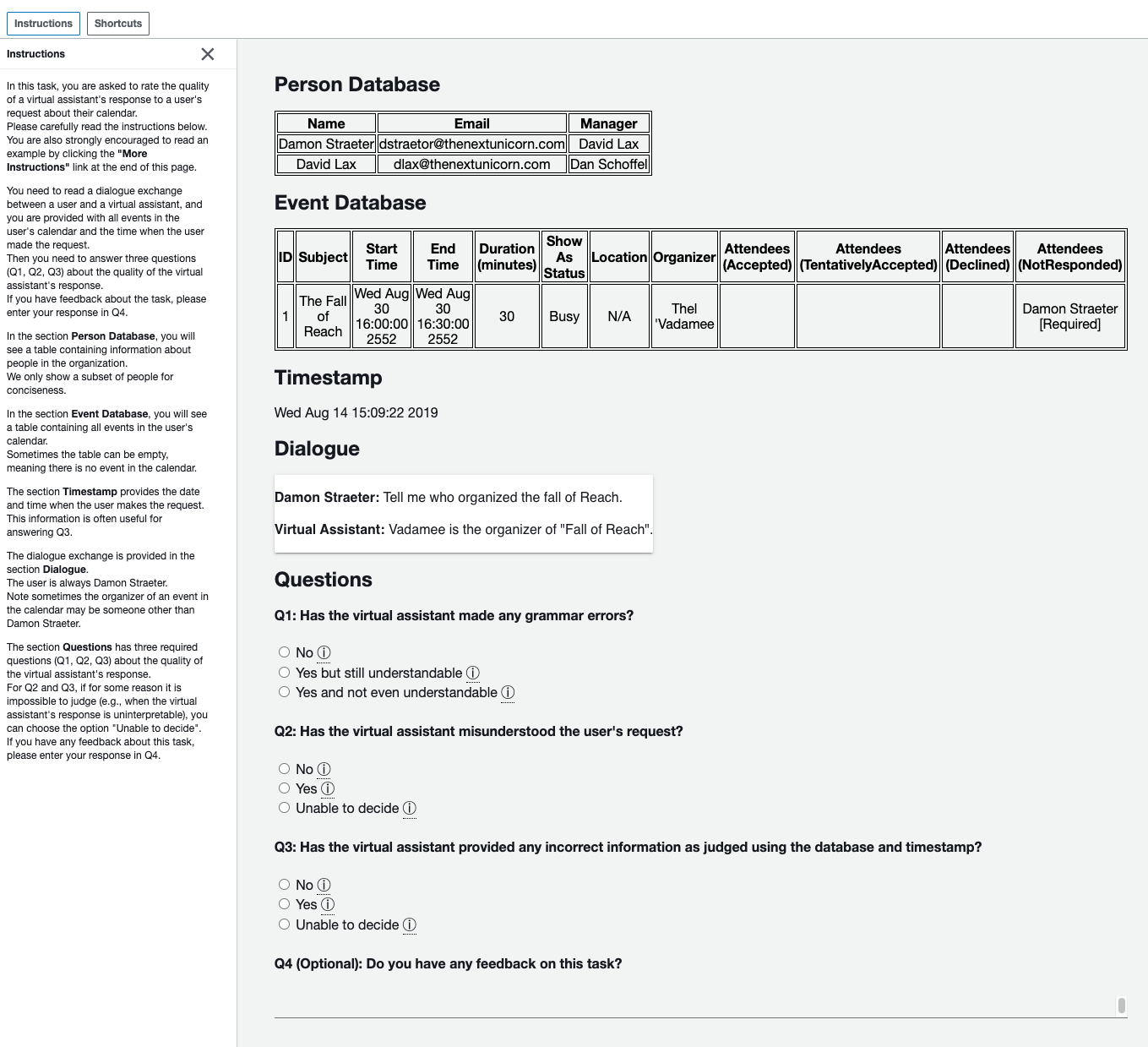}
    \caption{A screenshot of the MTurk interface for human evaluation.}
    \label{fig:mturk_interface}
\end{figure*}

\begin{table*}[t]
    \centering
    \footnotesize
    \begin{tabular}{l|ccc}
        \toprule
        \textbf{System} & \textbf{Grammatical} & \textbf{Relevant} & \textbf{Truthful} \\
        \midrule

        QCFG Random Sampling 
        & $.58$ & $.75$ & $.71$ \\

        Unconstrained Decoding 
        & $.86$ & $.71$ & $.71$ \\

        QCFG-Constrained Decoding
        & $.90$ & $.78$ & $.76$ \\

        Gold &
        $.95$ & $.81$ & $.80$ \\

        \bottomrule
    \end{tabular}
    \caption{The percentage of examples where all three workers choose the same answer.}
    \label{tab:inter_annotator_agreement}
\end{table*}

A screenshot of the MTurk interface for human evaluation is shown in \cref{fig:mturk_interface}.
\cref{tab:inter_annotator_agreement} shows the percentages of examples where all three workers choose the same answer for individual systems.
It can be observed that the gold responses receive the highest agreements on all three questions.
The QCFG-constrained decoding has slightly higher agreements than the unconstrained dcoding.
The QCFG random sampling receives a significantly lower agreement on ``Grammatical,'' probably because
this approach may produce ungrammatical responses but people may not agree on whether these are understandable.

\section{Model Configurations}
\label{appendix:hyperparams}

For SMCalFlow, we fine-tune the CodeT5 model for a fixed number of epochs (=10).
For MultiWOZ, we fine-tune the model for at most 10 epochs and do early stopping based on the on the loss on the development set.
We use the AdamW optimizer \cite{loshchilov-etal-2019-decoupled} with $\beta_1=0.9$ and $\beta_2=0.999$, 
using a linear learning rate scheduler with an initial learning rate of $5 \times 10^{-5}$.
For decoding, we always use a fixed beam size of $5$.

The CodeT5-base models used in our experiments have 220 million parameters. We used machines with 32GB V100 GPUs for model fine-tuning while the decoding experiments were carried out on CPU-only machines.

For SMCalFlow experiments, the input sequence to the LM is the string representation of the computation 
in the lispress format followed by its execution result rendered as a JSON string, e.g.,
``\textit{Plan: (Yield (Event.start ( $\ldots$ ))) Result: \{``type'': ``DateTime'', ``value'': $\ldots$ \} <s>}'',
where the last token is a special token to separate the input and the output.
For the ablative study (group 5) in \cref{ssec:ablation}, the user utterance is prefixed to the sequence, e.g.,
``\textit{User: When do I have thee oil change on my car scheduled for? Plan: $\ldots$ Result: $\ldots$ <s>}''.

For MultiWOZ experiments, the computation is rendered as a raw JSON string that encodes the ground-truth belief state 
and the predicted system act.
There is no execution result for these computations. 

\section{Qualitative Analysis}
\label{appendix:qualitative_analysis}

\begin{table}[t]
    \centering
    \footnotesize
    \begin{tabular}{l|ccc}
        \toprule
        & \textbf{Unconstrained} & \textbf{Constrained} \\
        \midrule

        Untruth
        & $19$ & $0$ \\

        Omission 
        & $3$ & $11$ \\

        Addition 
        & $17$ & $18$ \\

        Minor Difference
        & $10$ & $13$ \\

        Disfluency &
        $1$ & $1$ \\

        Annotation Error &
        $7$ & $8$ \\

        \midrule
        Total
        & 57 & 51 \\

        \bottomrule
    \end{tabular}
    \caption{Classification of differences between generated responses and human-annotated gold responses on 100 randomly sampled examples from
    the SMCalFlow dataset.
    Details are provided in \cref{appendix:qualitative_analysis}.}
    \label{tab:calflow_analysis}
\end{table}

We looked at 100 randomly selected examples from the experiments on SMCalFlow from \cref{ssec:main_results},
and compared the generated responses from both unconstrained decoding and QCFG-constrained decoding 
with the human-annotated gold responses provided by the dataset.
We summarize the differences between the generated and gold
responses in \cref{tab:calflow_analysis}, using the following categories:
\begin{description}
    \item[Untruth] The system reports incorrect information.
    \item[Omission] The system fails to mention information mentioned in the gold response.  
    \item[Addition] The system mentions additional (correct) information that is not mentioned in the gold response.
    \item[Minor Difference] The system uses a different phrasing than the gold response that nonetheless has the same information and fluency.
    \item[Disfluency] The system output is disfluent.
    \item[Annotation Error] The system output is acceptable but the gold annotation contains a fluency or factuality error.
\end{description}

For unconstrained decoding, 57 out of 100 responses differ from the gold responses,
whereas for QCFG-constrained decoding, only 51 of 100 responses differ.
This result is consistent with the R@1 column of
\cref{tab:calflow_main_results} (mismatch rates of 53\% and 44\%
respectively on the full validation set).

As expected, the most noticeable difference is in the number of Untruths.  The QCFG-constrained system produced no Untruths.  The unconstrained system produced 19\%, close to the 18\% rate found in the human evaluations in \cref{tab:calflow_main_results}.  
We show some examples of Untruths in \cref{tab:examples}. 

Conversely, the QCFG-constrained system produces substantially more Omissions than the unconstrained system.
Of the 11 omissions produced by the constrained system, 3 are are identical to the unconstrained output while 7 are on inputs for which the unconstrained output produce an Untruth.
In other words, our system successfully removed the 19 Untruths by the system, but in 7 of those cases, it produced a shorter (but still factually correct) input than the preferred gold annotation for that example.
We also note that the gold dataset is not consistent in how much information is included in the responses -- short answers like \textit{``Looks like it''} in Example C from \cref{fig:examples} are present in the gold annotations on examples similar to Example C.
Furthermore, both systems produce more Additions than Omissions, indicating that there is not a systematic bias towards shorter answers overall.
In future work, the model could be made to select more descriptive responses by adding a brevity penalty in the decoder or by weighting the QCFG productions, so that responses are scored not only by the LM but also by the QCFG.

\section{Dataset License}

The SMCalFlow dataset is distributed under the CC BY-SA 4.0 license.
To the best of the authors' knowledge, the MultiWOZ datasets were released under MIT license
as shown in \url{https://github.com/budzianowski/multiwoz}.
Our experiments follow the intended use of these datasets, which is to advance research in
dialogue systems.

\end{document}